\documentclass[letterpaper]{article} 
\usepackage{aaai25}  
\usepackage{times}  
\usepackage{helvet}  
\usepackage{courier}  
\usepackage[hyphens]{url}  
\usepackage[pagebackref,breaklinks,colorlinks]{hyperref}
\usepackage{graphicx} 
\usepackage{natbib}  
\usepackage{caption} 

\usepackage{algorithm}
\usepackage{algorithmic}

\usepackage{soul}
\usepackage{booktabs}
\usepackage{multirow}
\usepackage{amsmath}
\usepackage{amssymb}
\usepackage{cleveref}
\usepackage{bbding}
\usepackage{diagbox}
\usepackage{soul}
\usepackage{colortbl}
\usepackage{circledsteps}

\usepackage{newfloat}
\usepackage{listings}
\DeclareCaptionStyle{ruled}{labelfont=normalfont,labelsep=colon,strut=off} 
\floatstyle{ruled}
\newfloat{listing}{tb}{lst}{}
\floatname{listing}{Listing}
\title{UniMERNet: A Universal Network for \\ Real-World Mathematical Expression Recognition}
\author{
   Bin Wang$^{* 1}$, 
   Zhuangcheng Gu$^{* 1}$, 
   Guang Liang$^{* 1}$, \\
   Chao Xu $^{ 1}$,
   Bo Zhang $^{ 1}$,
   Botian Shi $^{ 1}$,
   Conghui He$^{\dag}$
}
\affiliations{
    \textsuperscript{\rm 1}Shanghai AI Laboratory, 
}

\usepackage{bibentry}

\begin{document}

\maketitle

\begin{abstract}
The paper introduces the UniMER dataset, marking the first study on Mathematical Expression Recognition (MER) targeting complex real-world scenarios. The UniMER dataset includes a large-scale training set, UniMER-1M, which offers unprecedented scale and diversity with one million training instances to train high-quality, robust models. Additionally, UniMER features a meticulously designed, diverse test set, UniMER-Test, which covers a variety of formula distributions found in real-world scenarios, providing a more comprehensive and fair evaluation. To better utilize the UniMER dataset, the paper proposes a Universal Mathematical Expression Recognition Network (UniMERNet), tailored to the characteristics of formula recognition. UniMERNet consists of a carefully designed encoder that incorporates detail-aware and local context features, and an optimized decoder for accelerated performance. Extensive experiments conducted using the UniMER-1M dataset and UniMERNet demonstrate that training on the large-scale UniMER-1M dataset can produce a more generalizable formula recognition model, significantly outperforming all previous datasets. Furthermore, the introduction of UniMERNet enhances the model's performance in formula recognition, achieving higher accuracy and speeds. All data, models, and code are available at \url{https://github.com/opendatalab/UniMERNet}.

\end{abstract}

\section{Introduction}
\label{sec:intro}

Mathematical Expression Recognition (MER) is a critical task in document analysis, aiming to convert image-based mathematical expressions into corresponding markup languages such as LaTeX or Markdown. MER is essential in applications like scientific document extraction, where a robust MER model helps maintain the logical coherence of documents. Unlike typical Optical Character Recognition (OCR) tasks, MER requires a deeper understanding of complex structures, including superscripts, subscripts, and various special symbols.

Existing research has primarily focused on enhancing the recognition accuracy of relatively simple rendered expressions~\cite{deng2017image} and handwritten data~\cite{mahdavi2019icdar,le2019pattern,wu2020handwritten,zhao2021handwritten} through a series of MER algorithms. Some researchers have begun to optimize MER algorithms by scaling up the training data and integrating them with transformer models~\cite{vaswani2017attention}, ensuring their applicability in diverse scenarios~\cite{kim2022ocr,pix2tex2022,blecher2023nougat,texify2023}. Other researchers have attempted to directly employ Large Vision-Language Models (LVLMs) for document content extraction, including MER~\cite{wei2023vary,blecher2023nougat}. However, existing MER benchmarks~\cite{deng2017image,mahdavi2019icdar} primarily focus on simple printed or handwritten expressions. Consequently, these models often struggle with diverse real-world expressions, such as lengthy equations and noisy scanned document screenshots.

\begin{figure}[t]
  \centering
	\includegraphics[width=1.0 \linewidth]{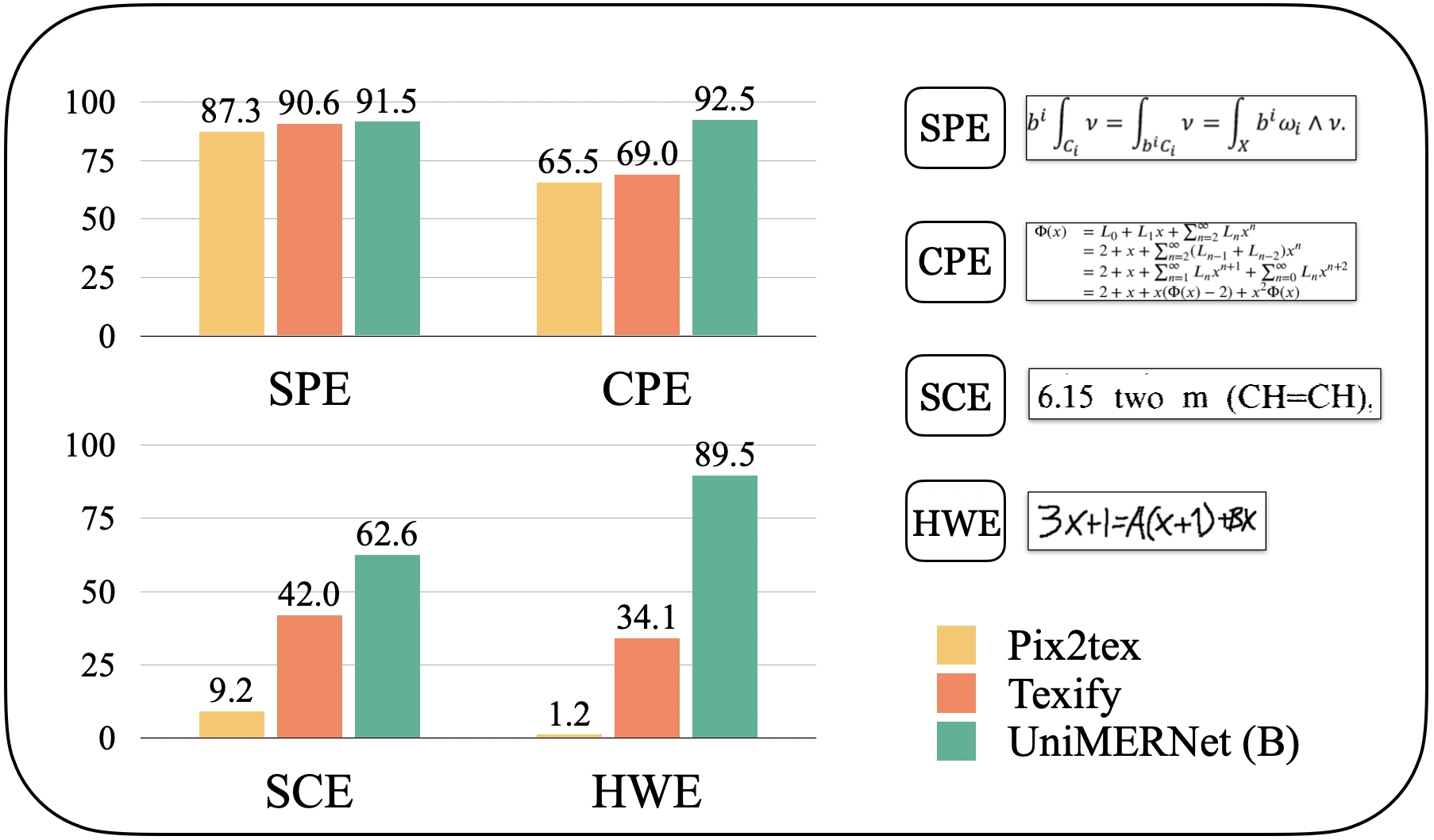}
    \caption{Performance comparison (BLEU Score) of mainstream models and UniMERNet in recognizing real-world mathematical expressions: Evaluation across Simple Printed Expressions (SPE), Complex Printed Expressions (CPE), Screen-Captured Expressions (SCE), and Handwritten Expressions (HWE).}
  \label{fig:fig1_introduction}
  \vspace{-12pt}
\end{figure}

In practice, real-world scenarios require the handling of complex, long expressions and noisy, distorted images from scanned documents or webpage screenshots.
To fill this gap, we introduce a comprehensive benchmark, UniMER-Test, which extends the existing test set with longer and real-world scenario expressions. 
Our benchmark aims to stimulate progress in MER by focusing on robustness and practical usage. As depicted in \Cref{fig:fig1_introduction}, we conduct exhaustive evaluations of state-of-the-art MER methods~\cite{pix2tex2022,texify2023} using our novel benchmark, UniMER-Test. These methods demonstrate remarkable competence in recognizing simple printed expressions. However, their performance noticeably declines when tested with more complex printed expressions, particularly long formulas.  The performance degradation becomes even more pronounced when these methods are applied to real-world expressions, such as screen-captured expressions embedded in noisy backgrounds and handwritten expressions. 
Moreover, large vision-language models such as Nougat~\cite{blecher2023nougat} and Vary~\cite{wei2023vary}, despite their capacity for convenient end-to-end document content extraction, exhibit only mediocre performance in MER.

To train a high-quality formula recognition model capable of accurately predicting results in diverse scenarios, we have constructed the UniMER-1M dataset. This large-scale dataset is specifically designed for Mathematical Expression Recognition (MER) and includes over one million diverse formula Image-LaTeX pairs. During its construction, we considered various levels of formula complexity, ranging from simple to complex long formulas, as well as different types, including printed and handwritten formulas. This ensures the dataset's suitability for training a model that generalizes well to real-world scenarios.  Furthermore, to fully leverage the UniMER dataset, we propose an innovative formula recognition model—UniMERNet. Unlike the mainstream document recognition frameworks that directly use the Swin-Transformer encoder and mBART decoder, we have optimized the model structure specifically for the formula recognition task. In the encoder, we introduce the Fine-Grained Embedding (FGE) module and the Convolutional Enhancement (CE) module for local context awareness. In the decoder, we incorporate the Squeeze Attention (SA) module to accelerate inference. These enhancements result in significant improvements in both inference speed and accuracy.

The main contributions of this paper are as follows:
\begin{itemize}
\item We introduce \textbf{UniMER}\footnote{\url{https://opendatalab.com/OpenDataLab/UniMER-Dataset}}~\cite{he2024opendatalab}, a universal MER dataset, with the training set UniMER-1M and the test set UniMER-Test, which encompasses all types of expressions in practical situations, offering a diverse and comprehensive foundation for MER model development and evaluation.
\item  We propose a novel network structure, \textbf{UniMERNet}, specifically designed for the formula recognition task. By designing a more precise encoder and a faster decoder, we can freely combine models to achieve higher accuracy and faster speed in formula recognition.

\item Validation of UniMERNet's superior performance through extensive experiments, establishing it as the new benchmark in open-source MER solutions by outperforming existing models in a variety of scenarios.

\end{itemize}

\section{Related Work}
\subsection{Traditional Machine Learning Methods in MER}

Decades ago, researchers recognized the importance of Mathematical Expression Recognition (MER). Anderson~\cite{anderson1967syntax} pioneered MER in irregular documents by introducing a parsing algorithm for two-dimensional character configurations. Miller and Viola~\cite{Miller_Viola_1998} proposed a system integrating character segmentation with the grammar of mathematical layouts. Chan \textit{et al.}~\cite{Chan_Yeung_1999} developed an online MER system featuring error detection and correction mechanisms. INFTY~\cite{suzuki2003infty} presented an OCR system for mathematical documents that achieved high character recognition accuracy through novel techniques. However, despite these advancements, MER precision was limited by handcrafted features in traditional machine learning.

\subsection{Deep Learning and Transformer Methods in MER}
With the advent of deep learning, various MER algorithms based on Convolutional Neural Networks (CNN)\cite{krizhevsky2012imagenet,simonyan2015very} were proposed. Deng \textit{et al.}\cite{deng2017image} introduced an encoder-decoder model with a coarse-to-fine attention mechanism, demonstrating superior performance over traditional OCR systems using the IM2LATEX-100K dataset. The WAP model~\cite{zhang2017watch} autonomously learned mathematical grammar and symbol segmentation, aligning closely with human intuition, while the PAL-v2 model~\cite{wu2020handwritten} used paired adversarial learning to excel in handwritten expression recognition on the CROHME dataset. Zhang \textit{et al.}\cite{zhang2020tree} proposed a tree-structured decoder for complex markups, and Zhao \textit{et al.}\cite{zhao2021handwritten} and Bian \textit{et al.}\cite{bian2022handwritten} enhanced MER with bi-directional learning in encoder-decoder models, advancing Handwritten Mathematical Expression Recognition (HMER). The CAN model\cite{li2022counting} improved HMER by incorporating a weakly supervised counting module, while Le \textit{et al.}\cite{le2019pattern} and Li \textit{et al.}\cite{li2020improving} employed data augmentation strategies to enhance MER performance.

More recently, the rapid development of Transformer models~\cite{vaswani2017attention} and large vision-language models~\cite{zhu2023minigpt,liu2024visual,dong2024internlm,liu2023improved,wang2024vigc,zhang2024internlm,chen2024far} led researchers to explore document information extraction task based on meticulously constructed evaluation benchmarks, such as DocGenome~\citep{xia2024docgenome} and MMSci~\citep{li2024mmsci}. For example, Donut~\cite{kim2022ocr} introduced an end-to-end model that converts document images into structured outputs without relying on OCR, while Nougat~\cite{blecher2023nougat} utilized auto-generated image-to-markup samples to train a Transformer-based encoder-decoder model. Vary~\cite{wei2023vary} offered a fine-grained multimodal model for document parsing. However, these methods often overlook the unique characteristics of mathematical expressions, leading to limitations in their MER capabilities. To address this, Pix2tex~\cite{pix2tex2022} and Texify~\cite{texify2023} trained encoder-decoder models on rendered mathematical expressions, though they struggle with complex or noisy expressions. 

In response to these challenges, the UniMERNet model proposed in this paper aims to build a robust and practical MER model that not only achieves state-of-the-art performance but also optimizes inference speed, enhancing the model’s applicability in real-world scenarios.

 \begin{table*}[t]
    \renewcommand\arraystretch{0.5}
    \centering

    \begin{tabular}{l|c|c|c|c|c}
    \toprule[1.0pt] 
    \textbf{Dataset}   & \textbf{ Type }& \textbf{ Train Size }&  \textbf{ Test Size } & \textbf{ Max Len } & \textbf{ Avg Len }  \\ \midrule[1pt]
    HME100K  & \multirow{2}{*}{HWE}         & 74,502    & 24,607    & 311  & 24.05  \\
    CROHME         &                        & 8,836     & 3233      & 147  & 22.27  \\  \midrule[0.5pt]
    IM2LATEX-100K  & \multirow{2}{*}{SPE}   & 83,883    & 10,354    & 440  & 96.01  \\  
    Pix2tex        &                        & 158,480   & 30,637    & 2949 & 93.35 \\  \midrule[0.5pt]
    UniMER     & Mixed                  & 1,061,791 & 23,757    & 7037 & 79.48  \\
    \bottomrule[1.0pt]
    \end{tabular}
    \vspace{-5pt}
        \caption{Statistical comparison of the MER dataset. ``Max Len'' and ``Avg Len'' mean the maximum length and average string length of the mathematical expression.}
    \label{tab:tab1_statistic}
    \vspace{-5pt}
\end{table*}

\begin{figure}[t]
  \centering
	\includegraphics[width=1 \linewidth]{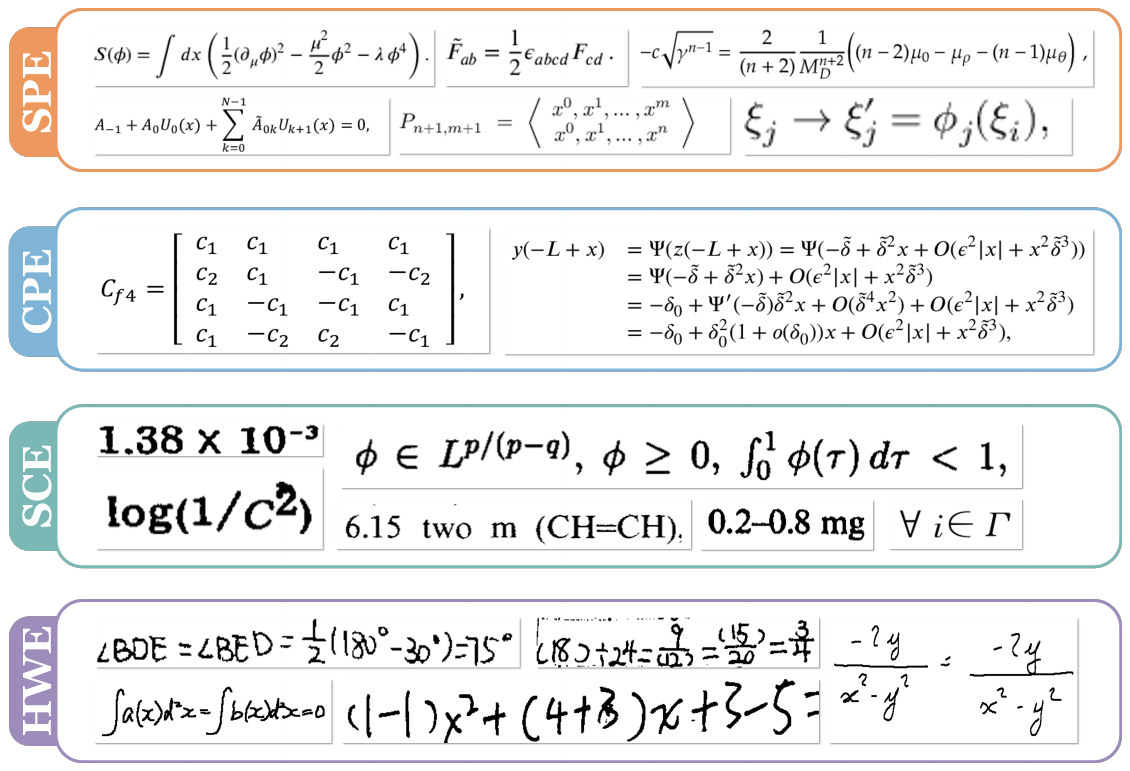}
    \caption{Visualization of the UniMER-Test dataset with four data types: Simple Printed Expressions (SPE), Complex Printed Expressions (CPE), Screen Capture Expressions (SCE), and Handwritten Expressions (HWE).}
  \label{fig:fig2_example}
  \vspace{-8pt}
\end{figure}

\section{UniMER Dataset}

The UniMER dataset addresses the diversity of formula recognition challenges in real-world scenarios. It consists of two main components: UniMER-1M, a large-scale training set, and UniMER-Test, a comprehensive evaluation set.

UniMER-1M includes 1,061,791 latex-image pairs, covering both simple and complex printed and handwritten formulas (\Cref{tab:tab1_statistic}). This extensive dataset surpasses existing formula recognition training sets, enabling the development of more robust models.

UniMER-Test, on the other hand, is a test set containing 23,789 samples. Unlike existing evaluation sets that primarily focus on simple printed and handwritten formulas, UniMER-Test comprehensively evaluates formula recognition across varying complexities and types, reflecting real-world scenarios (\Cref{fig:fig2_example}). Specifically, UniMER-Test includes the following types of formulas:

\begin{itemize}
    \item \textbf{SPE}: Formula images rendered from simple LaTeX expressions, characterized by uniform font size, clean background, and relatively short formulas.
    
    \item \textbf{CPE}: Formula images rendered from complex, long LaTeX expressions, characterized by uniform font size, clean background, and longer, more intricate formulas.
    
    \item \textbf{SCE}: Screen-captured images of formulas from documents and the web, characterized by inconsistent fonts and sizes, background noise, and image deformation.
    
    \item \textbf{HWE}: Collected from referenced handwriting recognition datasets~\cite{mouchere2014icfhr, mouchere2016icfhr2016, mahdavi2019icdar, yuan2022syntax}, these are complex and diverse, with varying backgrounds, but are relatively short.
\end{itemize}

This comprehensive approach ensures that UniMER-Test serves as a robust benchmark for evaluating formula recognition systems, setting a new standard for future research and development in the field.

\subsection{Data Collection Process}

\subsubsection{Printed Rendered Expressions (SPE, CPE)}

The assembly of our dataset begins with the Pix2tex~\cite{pix2tex2022} public dataset, which serves as the base for our SPE. Due to the limitations in volume and complexity, we expand the dataset by sourcing additional LaTeX expression codes from platforms like Arxiv, Wikipedia, and StackExchange. These codes are regularized~\cite{deng2017image} to resolve LaTeX syntax ambiguities, then compiled into expression PDFs in various fonts using XeLaTeX. Uncompilable expressions are discarded. Subsequently, ImageMagic's conversion function is utilized to transform these images into expressions with multiple DPIs, with data balancing ensuring an even distribution of different lengths.

Following this data expansion pipeline, we sample 725,246 simple formulas from the augmented data and combine them with the Pix2tex training set to form the SPE training data. The Pix2tex test set is designated as the SPE test data. In contrast, the CPE is derived independently of the Pix2tex dataset. We randomly select 110,332 complex formulas from the expanded data for training and test sets.


\begin{figure*}[th]
  \centering
	\includegraphics[width=0.92 \linewidth]{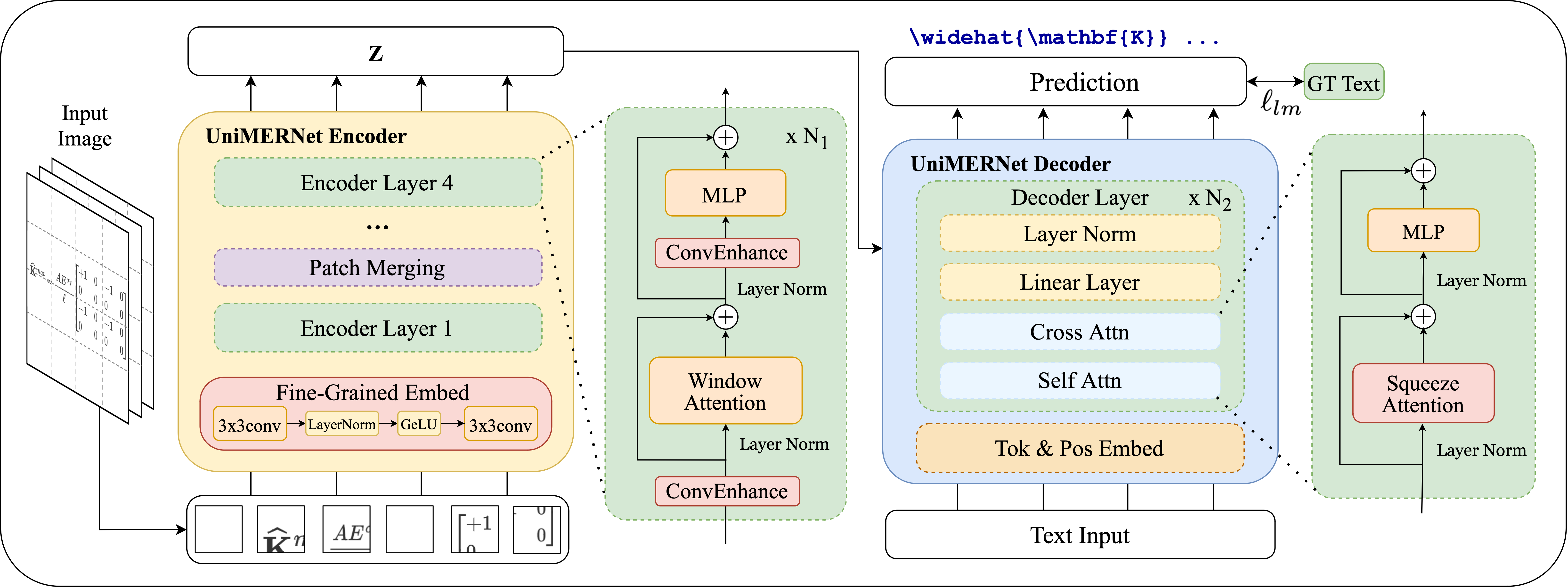}
    \caption{The overall framework of UniMERNet. The UniMER-Encoder incorporates Fine-Grained Embedding (FGE), Convolutional Enhancement (CE), and Removal of Shift Window (RSW) to enhance recognition capabilities. The UniMER-Decoder employs Squeeze Attention (SA) to accelerate inference speed.}
  \label{fig3:archtecture}
\end{figure*}

\begin{figure}[th]
  \centering
	\includegraphics[width=0.98 \linewidth]{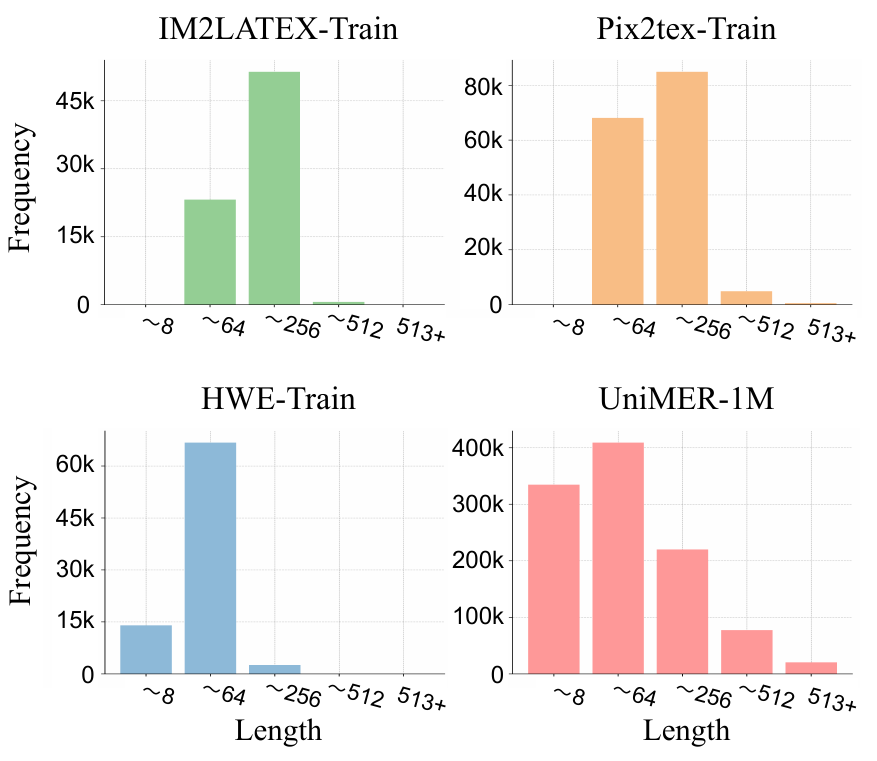}
    \caption{Formula string length distribution across datasets}
  \label{fig:fig4_distrib}
  \vspace{-8pt}
\end{figure}

\subsubsection{Screen-Captured Expressions (SCE)}

For SCE, we compile 1,000 diverse PDF pages in both Chinese and English, covering books, papers, textbooks, magazines, and newspapers. This variety ensures a wide range of fonts, sizes, and backgrounds for the formulas. Two annotators identify and label the formula boxes in the documents, capturing the content automatically. This process produces over 6,000 formula boxes, which are processed through Mathpix for formula recognition. After manual corrections and cross-verification by two annotators, redundant formulas are removed, resulting in 4,744 unique mathematical expression for the SCE test set.

\subsubsection{Handwritten Expressions (HWE)}

For HWE, we utilize the public datasets CROHME~\cite{mouchere2014icfhr,mouchere2016icfhr2016,mahdavi2019icdar} and HME100K~\cite{yuan2022syntax}. CROHME, a well-known dataset in HMER, originates from the handwritten digit recognition competition and includes 8,836 training expressions and 3,332 test expressions. HME100K, a real-world handwritten expression dataset, provides 74,502 training and 24,607 test images. Due to the high annotation accuracy of these datasets, we combine them for our HWE data. Specifically, the HWE training set consists of 8,836 formulas from CROHME and 74,502 from HME100K, totaling 83,338 samples. The HWE test set includes 3,332 formulas from CROHME and 3,000 from HME100K, totaling 6,332 test formulas.

\subsection{Diversified Training Data Sampling}

Existing formula datasets, such as HWE (CHROME \& HME100K)~\cite{mouchere2014icfhr,mouchere2016icfhr2016,mahdavi2019icdar}, IM2LATEX~\cite{deng2017image}, and Pix2tex~\cite{pix2tex2022}, primarily consist of rendered and handwritten formulas, but they have limitations in formula length and complexity. For example, Pix2tex mostly contains regular formulas, lacking extremely short or complex long formulas, while handwritten formulas are generally short with diverse styles, none exceeding 256 characters.

To address these limitations, we expand our UniMER-1M dataset with a wider range of formulas, sampled from sources like Arxiv and Wikipedia to ensure a balanced distribution of lengths and complexity. This varied sampling strategy enhances the model’s ability to recognize formulas across different complexity levels, improving overall performance. The formula length  distribution of IM2LATEX, Pix2tex, HWE, and UniMER-1M datasets is shown in ~\Cref{fig:fig4_distrib}.

\section{UniMERNet}

In real-world scenarios, mathematical formulas come from diverse sources such as electronic documents, scanned images, screenshots, and photographs. They range from single symbols to complex, lengthy expressions. Unlike general text recognition, formula recognition poses unique challenges in three dimensions. \textbf{Visual Similarity:} Many formula symbols look similar, e.g., \( \mu \) and \( u \), \( \beta \) and \( B \). This requires the model to have precise recognition capabilities. \textbf{Spatial Information:} Formulas often contain superscripts, subscripts, and other spatial arrangements, necessitating model's contextual awareness. \textbf{Inference Speed:} For complex and lengthy formulas, the symbol generation based on an encoder-decoder structure can be time-consuming, slowing down the model's inference speed.

To address these challenges, we design the UniMERNet formula recognition network. This network enhances recognition capabilities by incorporating fine-grained and context-aware modules in the encoder stage and accelerates inference speed by compressing the attention operation in the decoder stage.

Our architecture is based on the Swin-Transformer Encoder and mBART Decoder, which have been validated in various document processing tasks~\cite{kim2022ocr,blecher2023nougat,texify2023}. The overall framework of UniMERNet is shown in \Cref{fig3:archtecture}.

During training, each input formula image $\mathbf{I} \in \mathbb{R}^{3 \times H_0 \times W_0}$  undergoes an image augmentation module, transforming a single image representation into a diverse set of images. This effectively handles the varied representations of formulas in real-world scenarios. The UniMERNet encoder processes the image to generate a feature vector \( \mathbf{Z} \), which is then fed into the UniMERNet decoder. The decoder interacts with the feature vector \( \mathbf{Z} \) and the output text sequence via a cross-attention mechanism to generate the predicted formula. The decoder combines the feature vector \( \mathbf{Z} \), token embedding, and position embedding to predict the formula. For language modeling loss, we employ cross-entropy loss to minimize the difference between the predicted probability distribution of the next token and the actual distribution observed in the training data. The loss is defined as:
\begin{equation}
    \ell_{lm}(\hat{y}, y) = -\sum_{c=1}^{C} y_{o,c} \log(\hat{y}_{o,c}),
\end{equation}

Next, we detail the improvements in the UniMERNet network architecture tailored for formula recognition tasks.

\subsection{UniMERNet-Encoder}

\noindent \textbf{Fine-Grained Embedding (FGE).} 
The original Swin-Transformer uses a $4 \times 4$ single-layer convolutional kernel with a stride of 4 for patch embedding, resulting in a fourfold reduction in input resolution. However, these non-overlapping convolutional kernels often extract fragmented features, causing characters within formulas to be separated and adversely affecting recognition accuracy. Conversely, using larger overlapping convolutional kernels may lead to redundancy, as MER typically involves simpler elements like superscripts, subscripts, and various special characters, necessitating precise and streamlined feature extraction.

To address this, we implement an overlapping fine convolution layer composed of two convolutional layers, a Layer Normalization (LN) layer, and a GELU activation layer. Both convolutional layers utilize a kernel size of 3, a stride of 2, and padding of 1. This overlapping fine convolution expands the receptive field while mitigating fragmentation, thereby enhancing the model's overall performance.

\noindent\textbf{Convolutional Enhancement (CE).} While we retain the Window Attention mechanism from the Swin-Transformer, the receptive field within each window remains relatively large, limiting attention to small details such as labels in formulas and lacking inductive bias. Research in the visual domain ~\cite{guo2022cmt}~\cite{chu2021conditional}~\cite{d2021convit} suggests that convolutional and transformer models are complementary. In formula recognition, both global and local information are crucial. Global information enables the model to discern where to pause and where a new line begins, while local information helps the model understand the relationships between adjacent characters, particularly for local relationships like superscripts and subscripts.

Transformer architectures capture global information well, but discerning local details is critical. To address this, we introduce a local perception module called ConvEnhance, which enhances the model's ability to identify superscripts and subscripts. ConvEnhance, placed before each attention and MLP module, consists of 3x3 depthwise convolutions and GELU activation functions. The convolutions provide local perception, while the GELU activation implements gated threshold filtering. This addition allows UniMERNet to alternate between local and global information, significantly improving its performance.

\noindent \textbf{Removal of Shift Window (RSW).} The Shift Window operation aims to increase the model's receptive field by overlapping windows between layers, similar to convolutional kernels. However, with the introduction of FGE and CE, the receptive field is already significantly enlarged compared to the original Swin architecture. This makes the Shift Window operation redundant. Removing it not only improves the model's performance but also speeds up the model.

\subsection{UniMERNet-Decoder}

\noindent \textbf{Squeeze Attention (SA).} Experiments presented in the appendix indicate that the throughput bottleneck of UniMERNet lies within the language model, mBART. This is primarily because the visual model retrieves all visual features in a single pass, while the language model must iteratively utilize the previous prediction results to generate each successive token. Consequently, the throughput limitation is attributable to the language model. To address this, we introduce the concept of bottleneck from ResNet, implementing Squeeze Attention. Specifically, Squeeze Attention maps the query and key to a lower-dimensional space without excessive loss of information, thereby accelerating the computation of attention.

The enhanced UniMERNet-Encoder captures fine-grained and local contextual information, improving formula recognition accuracy. The optimized UniMER-Decoder, with slightly reduced precision, achieves significant speed improvements, making it highly effective for practical formula recognition systems.

\begin{table*}[t]
\footnotesize
\centering
\setlength{\tabcolsep}{4pt}
{
\begin{tabular}{lcccccccc}
\toprule[.9pt]
\midrule
\multirow{2}{*}{\begin{tabular}[c]{@{}l@{}}\textbf{Train}\\ \textbf{Dataset}\end{tabular}} & \multicolumn{2}{c}{\textbf{SPE}} & \multicolumn{2}{c}{\textbf{CPE}} & \multicolumn{2}{c}{\textbf{SCE}} & \multicolumn{2}{c}{\textbf{HWE}}  \\ 
\cmidrule(rl){2-3} \cmidrule(rl){4-5} \cmidrule(rl){6-7} \cmidrule(rl){8-9}  & BLEU $\uparrow$ & EditDis $\downarrow$& BLEU $\uparrow$ & EditDis $\downarrow$ & BLEU $\uparrow$ & EditDis $\downarrow$ & BLEU $\uparrow$ & EditDis $\downarrow$  \\  \midrule

Pix2tex    & 0.911 & 0.063 &  0.773 &  0.194  & 0.527 & 0.371 &  0.067 &  0.800   \\ 
Pix2tex\&HWE & 0.909 & 0.063 &  0.724 &  0.225  & 0.529 & 0.309 &  0.873 &  0.088   \\ 
UniMER-1M    & \textbf{0.915}  &  \textbf{0.060}&  \textbf{0.925 }&  \textbf{0.056 }&   \textbf{0.626}   &  \textbf{0.224} &  \textbf{0.895} &  \textbf{0.072}    \\
\bottomrule[.9pt]
\end{tabular}
\caption{Ablation results on UniMER-Test for models using different augmentations. Here, ``HME'' refers to a mixed dataset of CHROME and HME100K.}
\label{tab:tab2_dataset}
}
\end{table*}

\begin{table*}[thp]
\footnotesize
\centering
\setlength{\tabcolsep}{6.5pt}
{
\begin{tabular}{ccccccccccc}
\toprule[.9pt]
\multirow{2}{*}{\textbf{FGE}}& \multirow{2}{*}{\textbf{CE}} & \multirow{2}{*}{\textbf{RSW}} & \multirow{2}{*}{\textbf{SA}}  &
{\textbf{Params}} & {\textbf{FPS*}} && {\textbf{SPE}}  & {\textbf{CPE}} &{\textbf{SCE}} & {\textbf{HWE}}  \\ 
 \cmidrule(rl){5-11}  &&&&(M)&(img/s)&&BLEU $\uparrow$ &  BLEU $\uparrow$ & BLEU $\uparrow$ & BLEU $\uparrow$ \\  
\midrule
&&&&342&4.12&&  0.903  &  0.885 &  0.579 &  0.887 \\ 
\midrule

\multirow{1}{*}{\CheckmarkBold} &&&&342&4.10&& 0.903  &  0.888 &  0.584 &  0.886  \\ 
\midrule
\multirow{1}{*}{\CheckmarkBold} &\multirow{1}{*}{\CheckmarkBold}&&&342&4.07&& 0.912  &  0.896 &  0.599 &  0.895 \\
\midrule
\multirow{1}{*}{\CheckmarkBold} &\multirow{1}{*}{\CheckmarkBold}&&\multirow{1}{*}{\CheckmarkBold}&\textbf{325}&5.04&& 0.911  &  0.894 &  0.599&  0.893 \\ 
\midrule
\multirow{1}{*}{\CheckmarkBold} &\multirow{1}{*}{\CheckmarkBold}&\multirow{1}{*}{\CheckmarkBold}&\multirow{1}{*}{\CheckmarkBold}&\textbf{325}&\textbf{5.06}&& \textbf{0.912}  &  \textbf{0.897}  &  \textbf{0.601 }&  \textbf{0.893 }\\ 
\bottomrule[.9pt]
\end{tabular}
}
\caption{Ablation study of model architecture. \textbf{FGE}: replace with Fine-Grained Embedding, \textbf{CE}: add ConvEnhance module, \textbf{RSW}: Remove Shift Window, \textbf{SA}: apply Squeeze Attention in mBART decoder. \textbf{FPS*}: The model's throughput is tested on an A100 GPU with a batch size of 128, processing each sample up to a maximum sequence length of 1536.}
\label{tab:tab4_module}
\end{table*}

\section{Experiments}

\subsection{Datasets and Evaluation Metrics} 

We utilize the UniMER-1M dataset to train our model and evaluate its formula recognition performance using the UniMER-Test. Our evaluation relies on BLEU, Edit Distance, and ExpRate metrics.

\noindent\textbf{BLEU:} The BLEU score~\cite{papineni2002bleu}, initially developed for machine translation, quantifies the match of n-grams between candidate and reference sentences. Its application to a similar conversion task of formula recognition provides a robust, quantitative performance measure.

\noindent\textbf{Edit distance:} The Edit Distance~\cite{levenshtein1966binary} measures the minimum character changes needed to convert one string to another. Its use in formula recognition offers a precise, character-level accuracy assessment, making it a valuable performance metric.

\noindent\textbf{ExpRate:} Expression Recognition Rate~\cite{yuan2022syntax} is a widely used metric for handwritten formula recognition, defined as the percentage of predicted mathematical expressions that perfectly match the actual results.

\subsection{Implementation Details}
The proposed model, UniMERNet, uses PyTorch with a maximum sequence length set to 1536. Training is conducted on a single GPU equipped with CUDA. Specifically, we utilize an NVIDIA A100 with 80GB of memory. During the training phase, we employ eight such GPUs with a batch size of 64. The learning rate schedule is linear warmup cosine, with an initial learning rate of $1 \times 10^{-4}$, a minimum learning rate of $1 \times 10^{-8}$, and a warmup learning rate of $1 \times 10^{-5}$. Weight decay is set to 0.05. The total iteration is set to 300,000 by our default settings.

The architectural hyperparameters of UniMERNet instances are illustrated in ~\Cref{tab:instances}. Let $N$ denote the depth of the UniMERNet encoder, where [6, 6, 6, 6] indicates that each stage, from the first to the last, consists of six transformer layers.
Meanwhile, $M$ represents the depth of the UniMERNet Decoder. 
$C$ signifies the dimensionality of the vectors after processing through the Encoder, and 
Params refers to the total number of parameters in the model.
\begin{table}
		\begin{center}
        \resizebox{1.0\linewidth}{!}{
			\tiny
			\begin{tabular}{lcccccc}
				\hline
			   &$N$ & $M$  & $C$ &  Params    \\
				\hline
    UniMERNet-T    & [6, 6, 6, 6]     & 8     &   512   &  100M\\
    UniMERNet-S    & [6, 6, 6, 6]    & 8     &   768   &  202M\\
    UniMERNet-B    & [6, 6, 6, 6]    & 8     &   1024   &  325M\\
				\hline
			\end{tabular}
        }
\caption{Architectural hyper-parameters of UniMERNet. }
\label{tab:instances}
\end{center}
\end{table}

\subsection{Ablation Study}

\subsubsection{UniMER-1M} 

The diversity and quantity of training data are crucial for accurate formula recognition. As shown in \Cref{tab:tab2_dataset}, UniMERNet-B, trained solely on Pix2tex, achieves a BLEU score of 0.911 on SPE but performs poorly on CPE and HWE. The simplicity of Pix2tex leads to overfitting on SPE and difficulties with complex and handwritten formulas. Training on both Pix2tex and HWE improves the BLEU score on HWE to 0.873 but slightly declines on CPE. Notably, training with our UniMER-1M dataset, UniMERNet-B excels across all subsets. Compared to Pix2tex\&HWE, the CPE BLEU score improves by 0.201, and Edit Distance decreases from 0.225 to 0.056. On SCE, BLEU improves by 0.097, and Edit Distance decreases from 0.309 to 0.224. For HWE, BLEU improves by 0.022, and Edit Distance decreases to 0.072.

\begin{table*}[th]
\footnotesize
\centering
\setlength{\tabcolsep}{4.5pt}
{
\begin{tabular}{lcccccccccc}
\toprule[.9pt]
\midrule
\multirow{2}{*}{\textbf{Method}} & 
\multirow{2}{*}{\textbf{Params}} & 
\multirow{2}{*}{\textbf{FPS}} & \multicolumn{2}{c}{\textbf{SPE}} & \multicolumn{2}{c}{\textbf{CPE}} & \multicolumn{2}{c}{\textbf{SCE}} & \multicolumn{2}{c}{\textbf{HWE}}  \\ 
\cmidrule(rl){4-5} \cmidrule(rl){6-7} \cmidrule(rl){8-9} \cmidrule(rl){10-11} & (M)&(img/s) & BLEU $\uparrow$ & EditDis $\downarrow$& BLEU $\uparrow$ & EditDis $\downarrow$ & BLEU $\uparrow$ & EditDis $\downarrow$ & BLEU $\uparrow$ & EditDis $\downarrow$  \\  \midrule

Pix2tex~\cite{pix2tex2022}    &  -  &   -  &  0.873 &  0.088 &  0.655 &  0.408 & 0.092 &  0.817 &  0.012 &  0.920   \\ 
Texify~\cite{texify2023}     &  312  & 4.16 &  0.906 &  0.061 &  0.690 &  0.230 & 0.420 &  0.390 &  0.341 &  0.522    \\ 
\midrule

Texify*   & 312 & 4.16 & 0.906  &  0.067 &  0.900 &  0.077 & 0.599  & 0.224 & 0.888 & 0.075 \\ 

UniMERNet-T   &  107  &\textbf{7.20} & 0.909  &  0.066&  0.902 &  0.075 &  0.566  &  0.239 &  0.883 & 0.078 \\ 
UniMERNet-S   &  202  &6.04 & 0.913  &  0.061&  0.920 &  0.060 &  0.618  &  0.228 &  0.889 & 0.075 \\ 

UniMERNet-B   &  325  & 5.06 & \textbf{0.915}  &  \textbf{0.060}&  \textbf{0.925 }&  \textbf{0.056 }&   \textbf{0.626}   &  \textbf{0.224} &  \textbf{0.895} &  \textbf{0.072}  \\ 
\bottomrule[.9pt]
\end{tabular}
}
\vspace{-2pt}
\caption{Comparison with SOTA methods on UniMER-Test. \textit{Note:} Texify* is trained using UniMER-1M and the same data augmentation methods described in this paper.}\label{tab:tab_sota}

\end{table*}

\begin{table}[th]
\footnotesize
\centering
\setlength{\tabcolsep}{4pt}
{
\begin{tabular}{lcccccc}
\toprule[.9pt]
\midrule
\multirow{2}{*}{\textbf{Model}} &
\multirow{2}{*}{\textbf{Pre}} && {\textbf{SPE}}  & {\textbf{CPE}} &{\textbf{SCE}} & {\textbf{HWE}}  \\ 
 \cmidrule(rl){4-7} &&&BLEU $\uparrow$ &  BLEU $\uparrow$ & BLEU $\uparrow$ & BLEU $\uparrow$ \\  
 \midrule
Texify&
\multirow{1}{*}{\XSolidBrush}&& 0.903  &  0.884 &  0.576 &  0.886 \\
\midrule
Texify&
\multirow{1}{*}{\CheckmarkBold}&& 0.906  &  0.903 &  0.599 &  0.888 \\

 \midrule
UniMERNet-B&
\multirow{1}{*}{\XSolidBrush}&& 0.912 &  0.897 &  0.601 &  0.893\\
\midrule
UniMERNet-B&
\multirow{1}{*}{\CheckmarkBold}&& \textbf{0.915}  &  \textbf{0.925} &  \textbf{0.626} &  \textbf{0.895} \\

\bottomrule[.9pt]
\end{tabular}
}
\caption{Ablation of pre-training with text-image pairs. The \textbf{Pre} column indicates whether pre-training is used. Note: For Texify and UniMERNet-B, we use the same in-house text-image pre-training data for fair comparison.}
\vspace{-5pt}
\label{tab:tab_pretrain}
\end{table}

\subsubsection{Model Architecture Design} 

As shown in~\Cref{tab:tab4_module}, we conduct ablation experiments to validate our proposed optimization modules for the encoder, including Fine-Grained Embedding (FGE) and ConvEnhance (CN), and the decoder optimization module, Squeeze Attention (SA), along with the Remove Shift Window (RSW) operation. Our baseline architecture, \textbf{Texify}~\cite{texify2023}, uses the Swin-Transformer Encoder and mBART Decoder, similar to Donut~\cite{kim2022ocr} and Nougat~\cite{blecher2023nougat}. We train randomly initialized models from scratch using the UniMER-1M dataset.

From the comparison results, it is evident that incorporating the FGE and CE modules leads to a stable performance improvement across all subsets, with a significant increase observed in the SCE subset. The combination of these two modules improves the BLEU score from 0.579 to 0.599, an increase of nearly 2\%. When the SA module is applied to the decoder, the model's inference speed increases from 4.07 to 5.04, a relative improvement of 24\%, with minimal loss in accuracy. As mentioned in the methods section, with the CE module, the Shift Window becomes unnecessary. Removing the Shift Window results in slight improvements in both accuracy and speed, achieving the optimal configuration.

At this optimal configuration, the model's accuracy significantly improves compared to the baseline model, with BLEU score increases of 0.9\%, 1.2\%, 2.2\%, and 0.6\% on the SPE, CPE, SCE, and HWE subsets, respectively. The speed also improves by 23\%, demonstrating the substantial advantages of our UniMERNet in formula recognition.

\subsection{Pre-training with Text Image Pairs}

Pretraining on large-scale datasets significantly boosts model performance in document recognition. Models like Donut, Texify, and Nougat benefit greatly from this approach. Due to limited available data, we use Arxiv papers, applying text layout detection and OCR to extract text blocks and match them with source code, resulting in 16 million image-text pairs. As shown in \Cref{tab:tab_pretrain}, both the baseline model Texify and our UniMERNet-B exhibit significant improvements in SPE, CPE, and SCE metrics, with a relatively smaller gain in HWE. This is expected, as Arxiv papers predominantly feature printed text, differing from handwritten digit recognition. Incorporating more diverse pretraining data would further improve overall performance.

\subsubsection{Comparison with SOTA Methods}

To more intuitively evaluate the formula recognition performance of UniMERNet, we compared it with the state-of-the-art methods in the document recognition field. As shown in \Cref{tab:tab_sota}, when using the same network architecture as Texify, the baseline model Texify* significantly outperforms original Texify, underscoring the importance of the UniMER-1M dataset.  Moreover, our lightweight model UniMERNet-T, with an inference speed 1.73 times faster, already surpasses the previous SOTA model Texify. Using UniMERNet-B, the inference speed is 21.6\% faster compared to Texify. The BLEU scores on the SPE, CPE, SCE, and HWE subsets show absolute improvements of 0.009, 0.235, 0.206, and 0.554 respectively, demonstrating the superior recognition accuracy and robustness of our model across various scenarios.

\begin{figure}[t]
  \centering
	\includegraphics[width=1.\linewidth]{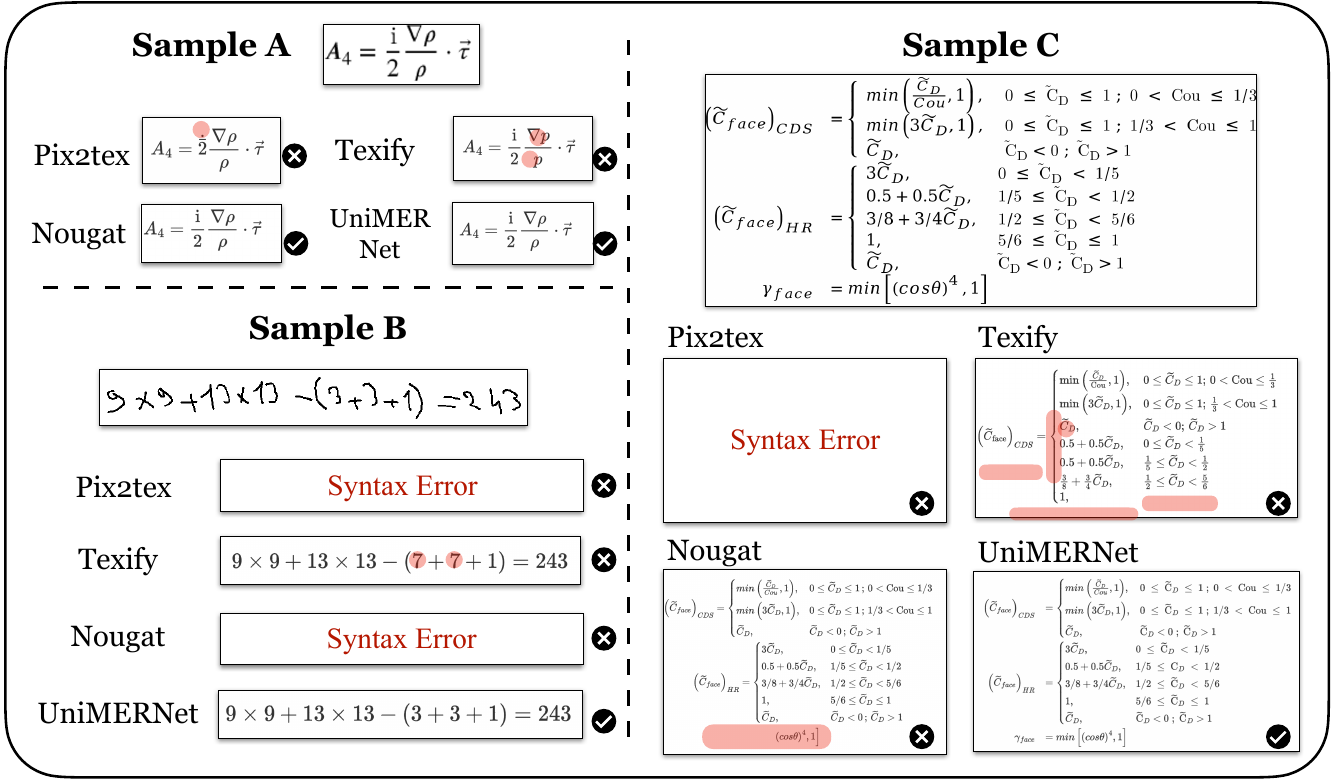}
    \caption{Comparative Visualization of Recognition Results Using Different Methods.}
  \label{fig:fig5_result}
  \vspace{-8pt}
\end{figure}

\subsubsection{Qualitative Comparisons}
As shown in \Cref{fig:fig5_result},  we selected three representative samples from the UniMER-Test set to thoroughly compare the performance between Pix2tex, Texify, Nougat, and UniMERNet. It's important to highlight that Nougat, being primarily designed for full-page recognition, tends to underperform with isolated formulas; thus, we prepared the test images by integrating random text with the formulas to adapt to Nougat's inference capabilities. Notably, while the other models exhibit certain shortcomings in handling these test samples, our model consistently delivers robust and accurate recognition results.

\section{Conclusion}
This paper introduces the large-scale training dataset UniMER-1M and the diverse evaluation dataset UniMER-Test, contributing significantly to robust formula recognition and fair evaluation. We also designe UniMERNet, a model with superior detail perception and contextual understanding, achieving high accuracy and speed, making it valuable for practical applications. Moving forward, we will explore using larger and more diverse pre-training data to enhance formula recognition. Additionally, we will investigate integrating UniMERNet with large vision-language models to improve the recognition of documents containing text, formulas, and tables, advancing document understanding.

\bigskip
\bibliography{unimernet} 

\clearpage
\appendix

\section*{Appendix}
\subsection{Details of UniMER Dataset}

\subsubsection{SPE and CPE Sampling} Existing datasets, such as IM2LATEX-100k and Pix2tex, present two primary challenges. Firstly, the size of these datasets, typically ranging from 100k to 200k formulas, is insufficient for training a precise and robust MER model. Secondly, these datasets contain a limited number of complex formulas, which compromises the model’s performance, particularly in handling multi-line complex expressions.

\begin{figure}[h]
  \centering
	\includegraphics[width=0.95 \linewidth]{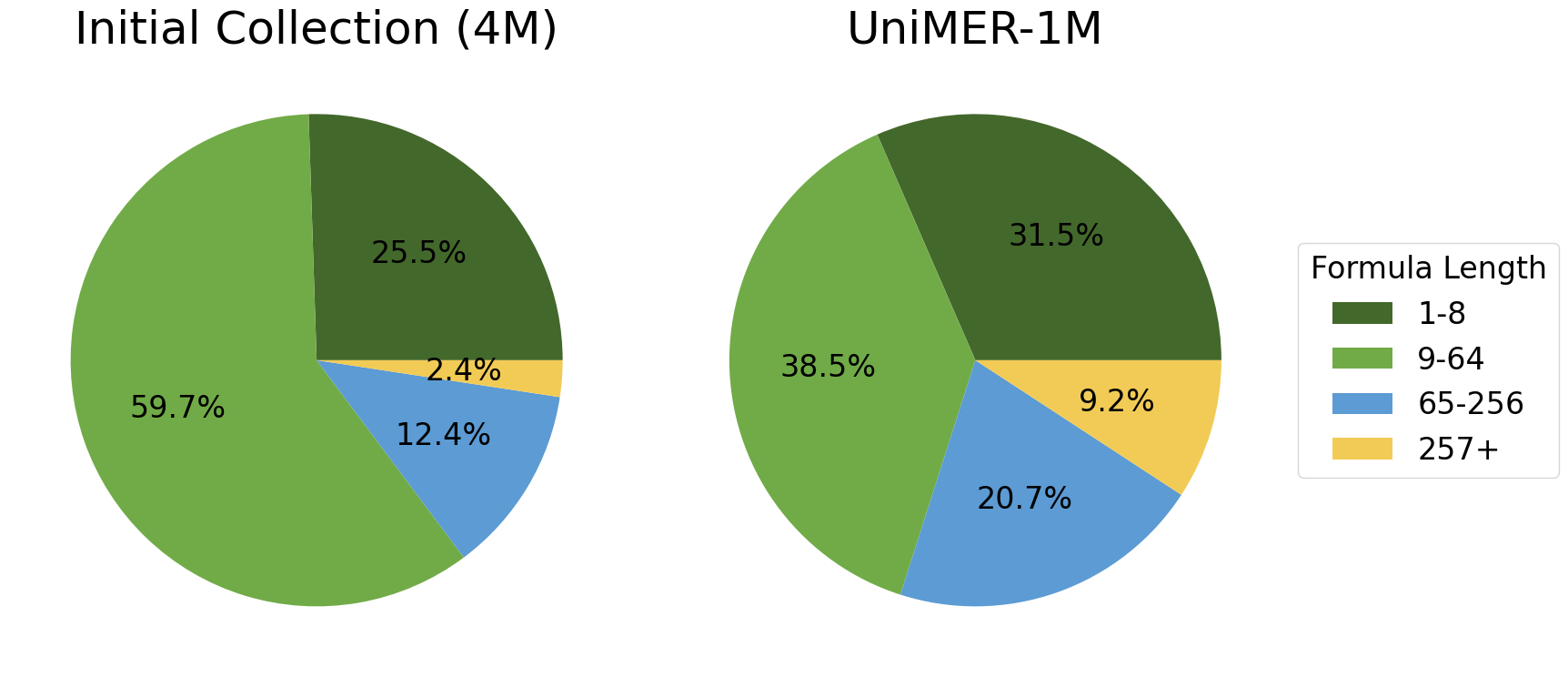}
    \caption{Formula length before and after re-sampling.}
  \label{fig:fig1_example}
\end{figure}

To address the limited size of the dataset, we expand it by incorporating an additional 4 million LaTeX expression source codes, building on the previously mentioned open-source datasets. These new entries are predominantly sourced from Arxiv (89\%), with supplementary contributions from Wikipedia (9\%) and StackExchange (2\%). This initial dataset expansion enhances the model’s overall capabilities. However, the proportion of long formulas in the initial collection is relatively small (2.3\%), which may cause inadequate training for complex expressions. To address this, we extract the longest formulas as CPE and adjust their ratio with randomly sampled SPE. This rearrangement ensures a balanced representation of varying lengths within the dataset, thereby significantly improving the model’s ability to recognize complex multi-line mathematical expressions. The distribution after rearrangement is shown in \Cref{fig:fig1_example}.

\subsubsection{SCE Deduplication} When extracting mathematical formulas from PDF pages, we face a unique challenge: formulas originating from the same page often appear identical in content, leading to potential duplicates. Simple deduplication based on textual content alone risks significant data loss, as identical formulas can appear across different pages, each bearing distinct visual characteristics such as font styles, sizes, and backgrounds. To preserve the richness of visual diversity while eliminating true duplicates, we adopted an image-based deduplication strategy, employing Perceptual Hashing to assess image similarity. This method allows us to compare the visual features of the formula images directly, ensuring that only those with high similarity—indicating true duplicates—are removed. Through this meticulous process of image similarity analysis, we effectively reduced the dataset to 4,744 unique Screen-Captured Expressions (SCE), each representing a distinct visual instance of mathematical expressions, thereby constituting our refined SCE test set.

\begin{lstlisting}[caption=XeLaTeX rendering setting, label={lst:example}, basicstyle=\scriptsize\ttfamily]
\documentclass[varwidth]{standalone}
\usepackage{fontspec,unicode-math}
\usepackage[active,displaymath,textmath,tightpage]{preview}
\usepackage[total={16in, 16in}]{geometry}
\setmathfont{
    % MATH_FONT
}
\begin{document}
\thispagestyle{empty}
\begin{displaymath}
    % MATH_FORMULA
\end{displaymath}
\end{document}
\end{lstlisting}

\subsubsection{Rendering Settings} For the rendering settings, we follow the similar procedure used in \cite{deng2017image} and \cite{pix2tex2022}. The dataset is rendered using XeLaTeX with a diverse range of math fonts and DPI settings. The chosen fonts included Asana Math, Cambria Math, XITS Math, GFS Neohellenic Math, TeX Gyre Bonum Math, TeX Gyre Dejavu Math, TeX Gyre Pagella Math, and Latin Modern Math, with Latin Modern Math as the default math font being employed in approximately 22\% of the cases. To accommodate different levels of clarity and detail, the DPI setting varies between 80 to 350 when converting to PNG format, allowing for adjustments in the resolution and sharpness of the rendered mathematical expressions. The rendering template we use is shown in Listing \ref{lst:example}.

\begin{figure}[t]
  \centering
	\includegraphics[width=0.95 \linewidth]{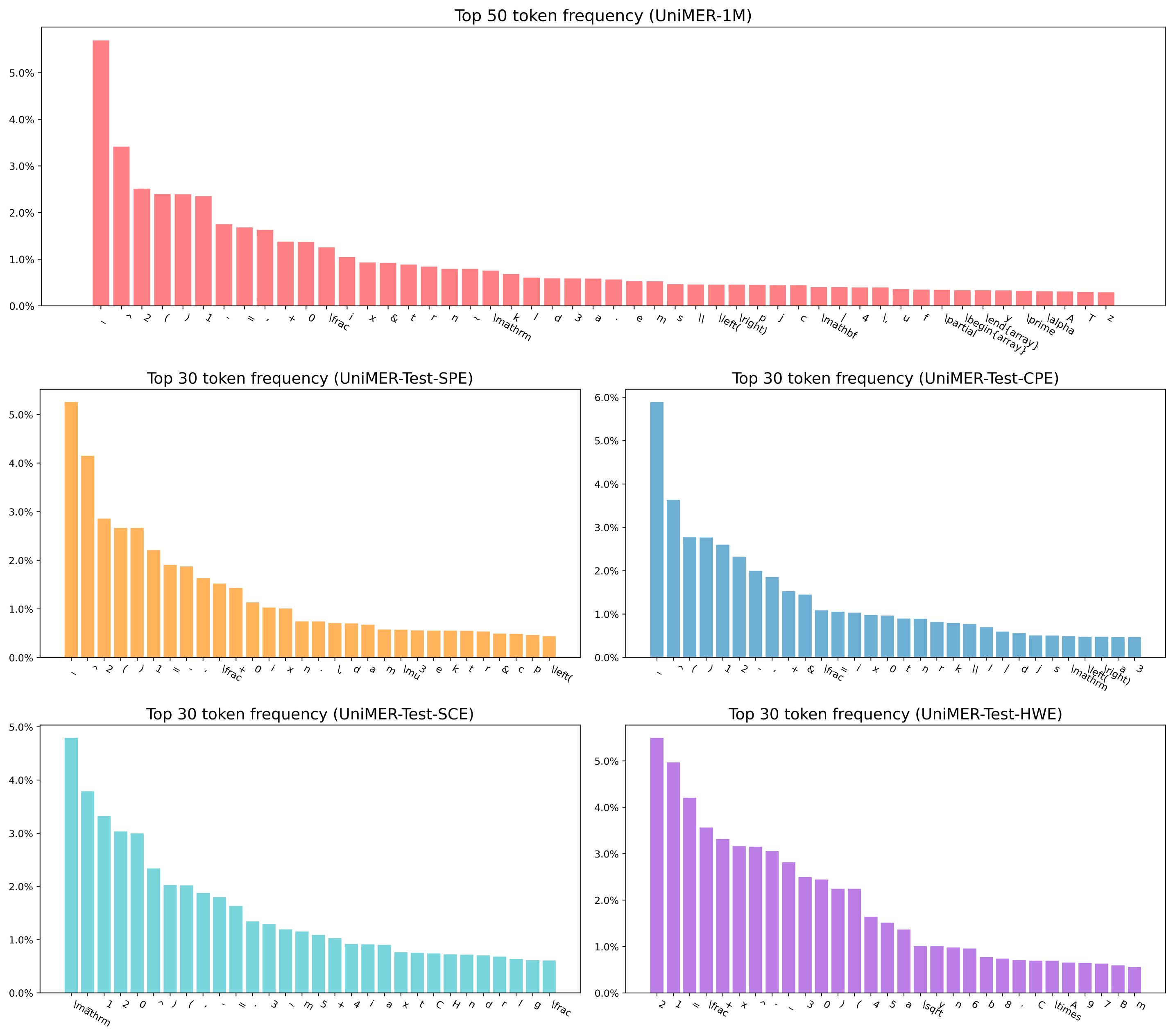}
    \caption{Most frequent occurring latex symbols in UniMER-1M and UniMER-Test subsets}
  \label{fig:fig2_tok}
\end{figure}

\subsubsection{Formula Text Normalization} LaTeX syntax inherently contains ambiguous information, as different source codes can produce the same rendering. This presents significant challenges in the evaluation phase of the math formula recognition task's benchmark because it potentially leads to incorrect assessments of a model's performance despite producing visually identical formula renderings. In handwritten math recognition datasets, such as CROHME, a self-defined label graphs format is used, eliminating ambiguous expressions by employing a character relation-based method. We do not adopt these methods for normalization as they involve format conversions during model training and, more importantly, use only partial LaTeX syntax.

The LaTeX normalization is first introduced in~\cite{deng2017image}. This preprocessing operation involves fixing super-script and sub-script order, replacing ambiguity with unified expressions while resulting in no or minimal visual changes in rendering, preserving the integrity of the original mathematical expressions. Subsequent datasets, such as IM2LATEX-100K~\cite{deng2017image} and Pix2tex~\cite{pix2tex2022}, have adopted similar methods. Building on this foundation of normalization, we have adjusted the normalization rules for certain LaTeX environments, enabling better support for multi-line formula expressions and previously unsupported syntax. All formulas in UniMER-1M and UniMER-Test undergo this normalization process to facilitate better horizontal comparison with previous datasets that employ a similar normalization process.

\subsection{Data Statistics}

\subsubsection{Most Occurring Symbols} Diving into the dataset's LaTeX symbols offers intriguing insights into the most frequently utilized mathematical notations. The bar chart provided in the \Cref{fig:fig2_tok} illustrates the frequency of specific LaTeX symbols that appear in UniMER. Symbols such as Greek letters, operators, and various mathematical functions are universally prevalent in each dataset, underlining their fundamental role in articulating complex mathematical ideas. A subtle variation is observed in the SCE and HWE datasets, where numbers and letters are noticeably more frequently occurring, as they contain relatively easier and less structured math expressions.

\subsubsection{Image Size Distribution}
\begin{figure}[t]
  \centering
	\includegraphics[width=0.8 \linewidth]{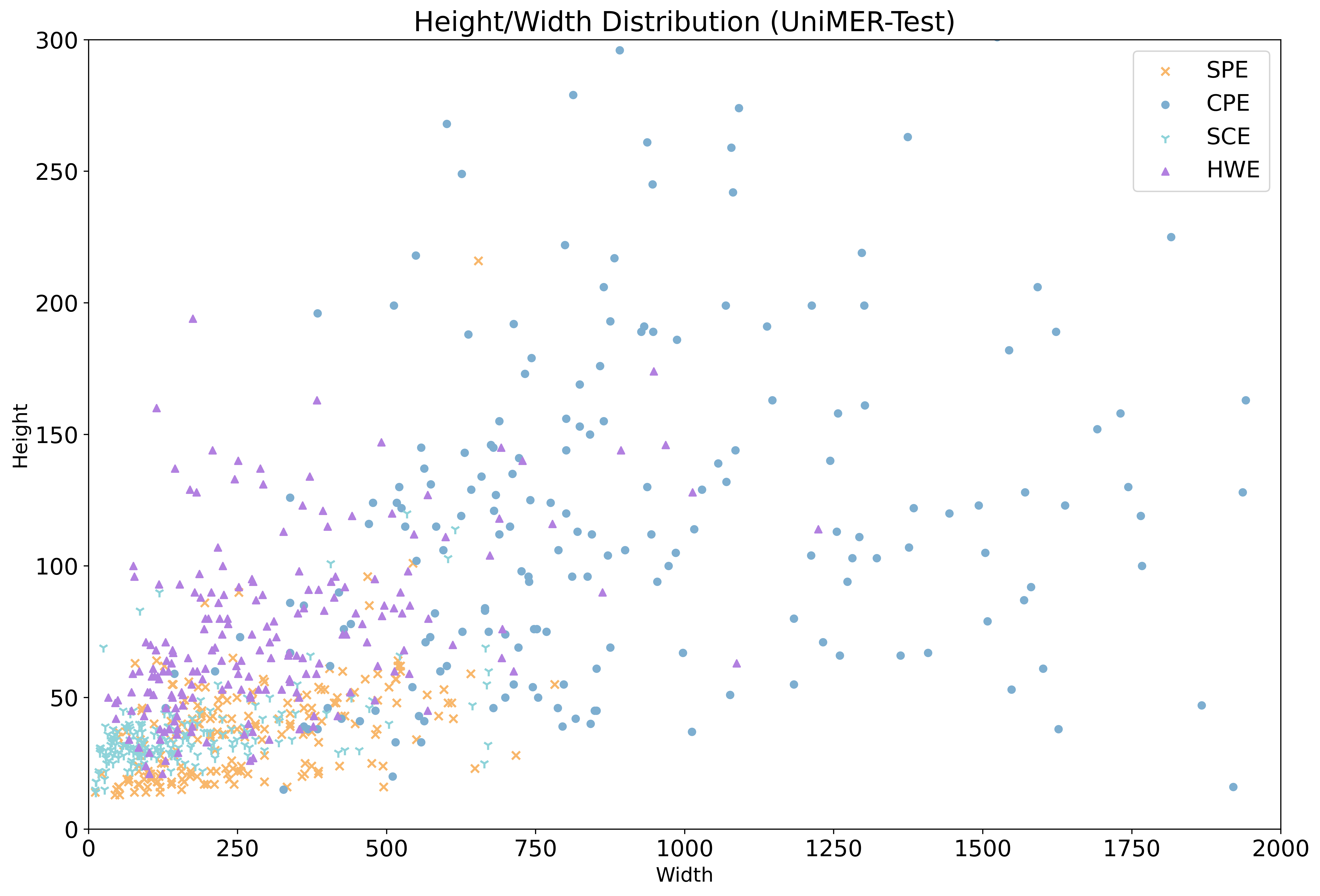}
    \caption{Height width scatter plot in UniMER-Test subsets with sampling}
  \label{fig:fig3_hw}
\end{figure}

The scatter plot in \Cref{fig:fig3_hw} provides a visual distribution of image sizes across different subsets within the UniMER-Test. Each point on the plot represents an individual image, with its position determined by the image's width and height. The SPE, CPE, SCE, and HWE subsets each exhibit unique clusters, indicating the variety in dimensionality they encompass. It's evident that the SPE and SCE subsets tend to have a higher concentration of smaller images, as shown by the dense clustering of points towards the lower end of the spectrum. The distribution of image sizes within the CPE dataset exhibits a considerable spread, highlighting the diversity of dimensions that this particular subset encompasses, indicating its complexity compared to SPE. On the other hand, the HWE subset is characterized by images with generally larger dimensions. This can be attributed to the fact that these images are often photographed and contain noise, necessitating a higher resolution to ensure that the finer details of the handwritten expressions are preserved and recognizable.

\begin{table*}[t]
 \footnotesize
\centering
\resizebox{0.95\linewidth}{!}{
\setlength{\tabcolsep}{4pt}
{
\begin{tabular}{lccccccccc}
\toprule[.9pt]
\midrule
\multirow{2}{*}{\begin{tabular}[c]{@{}l@{}}\textbf{Train}\\ \textbf{Dataset}\end{tabular}}& \multirow{2}{*}{\textbf{Augment}} & \multicolumn{2}{c}{\textbf{SPE}} & \multicolumn{2}{c}{\textbf{CPE}} & \multicolumn{2}{c}{\textbf{SCE}} & \multicolumn{2}{c}{\textbf{HWE}}  \\ 
\cmidrule(rl){3-4} \cmidrule(rl){5-6} \cmidrule(rl){7-8} \cmidrule(rl){9-10} &  & BLEU $\uparrow$ & EditDis $\downarrow$& BLEU $\uparrow$ & EditDis $\downarrow$ & BLEU $\uparrow$ & EditDis $\downarrow$ & BLEU $\uparrow$ & EditDis $\downarrow$  \\  \midrule
\multirow{2}{*}{Pix2tex}  
& \XSolidBrush   & 0.909  &  0.064 &  0.764 &  0.198 &   0.512 &  0.380 &  0.065 &  0.807    \\ 
& \CheckmarkBold & 0.911 & 0.063 &  0.773 &  0.194  & 0.527 & 0.371 &  0.067 &  0.800 \\ 
\midrule
\multirow{2}{*}{UniMER-1M} 
& \XSolidBrush  &   0.912 & 0.064 & 0.911 & 0.063 &   0.601 &  0.251 &  0.886 &  0.078    \\ 
& \CheckmarkBold  & \textbf{0.915}  &  \textbf{0.060}&  \textbf{0.925 }&  \textbf{0.056 }&   \textbf{0.626}   &  \textbf{0.224} &  \textbf{0.895} &  \textbf{0.072}   \\
\midrule
\bottomrule[.9pt]
\end{tabular}
}
\vspace{-3pt}
}
\caption{Ablation results on UniMER-Test with models using different augmentations.}
\label{tab:tab_dataAug}
\vspace{-5pt}
\end{table*}

\begin{figure*}[t]
  \centering
	\includegraphics[width=0.8 \linewidth]{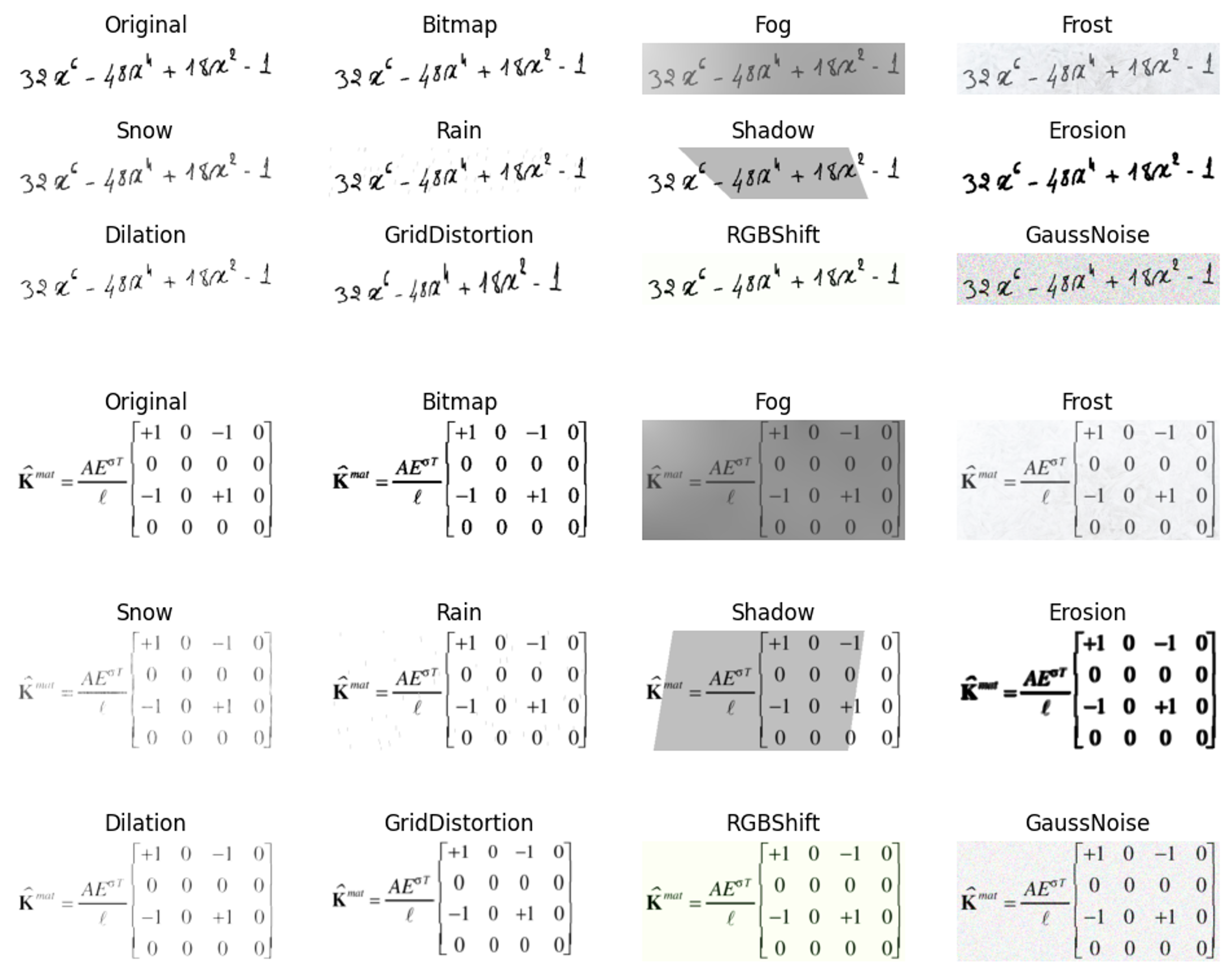}
    \caption{Visualization of selected image augmentations applied during training. } \label{fig:fig4_aug}
\end{figure*}

\subsection{Data Augmentation} 

While introducing additional training data in UniMER-1M enhances the variety of formulas, it does not account for the diversity of real-world formula images, which can come from scanned documents or photos and can exhibit noise and distortion. We employ various image augmentation techniques during model training to simulate this diversity with extra transformations from Albumentations~\footnote{https://albumentations.ai} library and self-defined transformations, which include but are not limited to:
\begin{itemize}
    \item \textbf{Erosion/Dilation} - To simulate the textural imperfections often found in screen-captured formulas, these operations modulate the thickness of characters, mirroring the effects of resolution differences and printer anomalies.
    \item \textbf{Degradation Simulation} (Fog, Frost, Rain, Snow, Shadow) - These augmentations introduce environmental artifacts to mimic the conditions under which documents might be photographed in real-world scenarios, adding layers of complexity such as blurriness and occlusions.
    \item \textbf{Geometric Transformations} (Rotation, Distortion …) - To account for the angle and perspective distortions typical in photographed or scanned documents, these operations adjust the orientation and shape of the mathematical expressions.
\end{itemize}
Each image undergoes a sequence of these augmentation operations with a given probability. This helps to bridge the gap between the pristine, synthesized training data and the noisy, real-world test images and improves UniMERNet's performance for real piratical use. \Cref{fig:fig4_aug} provides a visualization of selected transformations.

\subsubsection{Impact of Data Augmentation on Performance}

Data augmentation proves to be significantly beneficial for real-world formula recognition tasks. As shown in \Cref{tab:tab_dataAug}, incorporating image augmentation into the training process with the Pix2tex dataset leads to varying degrees of improvement across all evaluation subsets. Notably, on the SCE subset, the BLEU score increases by 2.5\%. Similar trends are observed when training with the UniMER-1M dataset, where the BLEU score on the SCE subset improves significantly from 0.601 to 0.626, and the edit distance decreases from 0.251 to 0.224,/;.

\section{Optimal Depth Configuration for UniMERNet} 
In this section, we investigate the optimal depth configuration for the encoder and decoder of the UniMERNet model. Through a series of comparative experiments, we analyze the impact of varying depths on model performance and throughput. 

\subsubsection{Encoder Depth Analysis}  Table \ref{tab:Arch1} presents the results of our experiments on different encoder depths while keeping the decoder depth constant. The findings indicate that increasing the encoder depth enhances model performance up to a certain point. Specifically, when the encoder depth reaches six layers per stage, the performance gains begin to plateau. This is evident from the BLEU scores across various evaluation metrics, which show diminishing returns beyond this depth. Additionally, the frames per second (FPS) metric reveals that increasing the encoder depth has a minimal impact on model throughput.

\begin{table*}[th]
\footnotesize
\centering
\setlength{\tabcolsep}{4pt}
{
\begin{tabular}{llccccccccc}
\toprule[.9pt]
\midrule
 \multirow{2}{*}{\textbf{N}} & \multirow{2}{*}{M}  &&{\textbf{Params}} & {\textbf{FPS}} &
{\textbf{SPE}}  & {\textbf{CPE}} &{\textbf{SCE}} & {\textbf{HWE}} & {\textbf{AVG}}  \\ 
 \cmidrule(rl){4-10} &&& (M)&(img/s)  &BLEU $\uparrow$ &  BLEU $\uparrow$ & BLEU $\uparrow$ & BLEU $\uparrow$ & BLEU $\uparrow$ \\  
 \midrule
$[2, 2, 2, 2]$     & 6 && 148  &7.25 & 0.890& 0.832& 0.496 & 0.833 & 0.763 \\ 
$[4, 4, 4, 4]$     & 6 && 167  &7.23 &0.897 ({+0.7\%}) & 0.856 ({+2.4\%})&0.544 ({+4.8\%}) &0.870 ({+3.7\%}) & 0.792 ({+2.9\%})\\ 
\cellcolor{gray!15}$[6, 6, 6, 6]$     \cellcolor{gray!15}& 6 \cellcolor{gray!15}&\cellcolor{gray!15}& \cellcolor{gray!15}186  &\cellcolor{gray!15}7.20 &\cellcolor{gray!15}0.901 ({+0.4\%})& \cellcolor{gray!15}0.878 ({+1.2\%}) &\cellcolor{gray!15}0.564  ({+2.0\%})&\cellcolor{gray!15}0.886 ({+1.6\%}) & \cellcolor{gray!15}0.807 ({+1.5\%})\\ 
$[8, 8, 8, 8]$     & 6 && 205  &7.15 &0.902 ({+0.1\%}) & 0.880 ({+0.2\%})& 0.578 ({+1.4\%})&0.878 ({-0.8\%}) & 0.809  ({+0.2\%})\\ 

\bottomrule[.9pt]
\end{tabular}
}
\caption{Comparative study of the impact of encoder depth on model performance. Let $N$ denote the depth of the UniMERNet encoder, where [6, 6, 6, 6] indicates that each stage, from the first to the last, consists of six transformer layers. Meanwhile, $M$ represents the depth of the UniMERNet Decoder.}
\label{tab:Arch1}
\end{table*}

\begin{table*}[th]
\footnotesize
\centering
\setlength{\tabcolsep}{4pt}
{
\begin{tabular}{llccccccccc}
\toprule[.9pt]
\midrule
 \multirow{2}{*}{\textbf{N}} & \multirow{2}{*}{M}  &&{\textbf{Params}} & {\textbf{FPS}} &
{\textbf{SPE}}  & {\textbf{CPE}} &{\textbf{SCE}} & {\textbf{HWE}} & {\textbf{AVG}}  \\ 
 \cmidrule(rl){4-10} &&& (M)&(img/s)  &BLEU $\uparrow$ &  BLEU $\uparrow$ & BLEU $\uparrow$ & BLEU $\uparrow$ & BLEU $\uparrow$ \\  
 \midrule
$[6, 6, 6, 6]$     & 4 && 169  &8.36 & 0.893& 0.850 & 0.532 & 0.863 & 0.785\\ 
$[6, 6, 6, 6]$     & 6 && 186  &7.20 &0.901({+0.8\%}) & 0.878({+1.8\%})&0.564({+3.2\%})&0.886({+2.3\%}) &0.807 ({+2.2\%})\\ 
\cellcolor{gray!15}$[6, 6, 6, 6]$     \cellcolor{gray!15}& 8 \cellcolor{gray!15}&\cellcolor{gray!15}& \cellcolor{gray!15}202  &\cellcolor{gray!15}6.04 &\cellcolor{gray!15}0.905({+0.4\%})&\cellcolor{gray!15}0.892({+1.6\%})&\cellcolor{gray!15}0.587({+2.3\%}) &\cellcolor{gray!15}0.888({+0.2\%})&\cellcolor{gray!15}0.818({+1.1\%})\\ 
$[6, 6, 6, 6]$     & 10 && 219  &4.99 & 0.907({+0.2\%}) & 0.894({+0.2\%})& 0.582({-0.5\%}) & 0.893({+0.5\%}) & 0.819({+0.1\%}) \\

\bottomrule[.9pt]
\end{tabular}
}
\caption{Comparative study of the impact of decoder depth on model performance. Let $N$ denote the depth of the UniMERNet encoder, where [6, 6, 6, 6] indicates that each stage, from the first to the last, consists of six transformer layers. Meanwhile, $M$ represents the depth of the UniMERNet Decoder.}
\label{tab:Arch2}
\end{table*}

\subsubsection{Decoder Depth Analysis} Table \ref{tab:Arch2} explores the effects of varying decoder depths while keeping the encoder depth fixed at six layers per stage. Similar to the encoder, increasing the decoder depth improves performance up to a depth of eight layers, after which the performance gains diminish significantly. However, unlike the encoder, the decoder depth has a more pronounced impact on throughput, with FPS decreasing notably as the decoder depth increases.

By applying the law of diminishing marginal utility and Pareto optimality principles, we determine that the optimal configuration for UniMERNet is an encoder depth of six layers per stage and a decoder depth of eight layers. This configuration balances model performance and throughput, ensuring efficient and effective processing.

\end{document}